\documentclass{llncs}

\usepackage{url}
\usepackage{array}
\usepackage{bbm}
\usepackage{diagbox}
\usepackage{tabularx}
\usepackage{graphicx}
\usepackage{color}
\usepackage{ragged2e}
\usepackage{amsmath}
\usepackage{amssymb}
\usepackage{booktabs}
\usepackage[colorlinks,
linkcolor=blue,
anchorcolor=red,
citecolor=green
]{hyperref}

\newif\ifshowcomments
\showcommentstrue

\ifshowcomments

\fi

\begin{document}
\mainmatter              

\title{Probabilistic Multilayer Regularization Network for Unsupervised 3D Brain Image Registration}

\titlerunning{}  
\author{Lihao Liu, Xiaowei Hu, Lei Zhu*, and Pheng-Ann Heng}
\authorrunning{Lihao Liu et al.} 
\institute{Dept. of Computer Science and Engineering, The Chinese University of Hong Kong}

\maketitle
\let\thefootnote\relax\footnotetext{* Corresponding author: \email{lzhu@cse.cuhk.edu.hk}.}

\begin{abstract}

Brain image registration transforms a pair of images into one system with the
matched imaging contents, which is of essential importance for brain image analysis.
This paper presents a novel framework for unsupervised 3D brain image registration by capturing the feature-level transformation relationships between the unaligned image and reference image.
To achieve this, we develop a feature-level probabilistic model to provide the direct regularization to the hidden layers of two deep convolutional neural networks, which are constructed from two input images.
This model design is developed into multiple layers of these two networks to capture the transformation relationships at different levels.
We employ two common benchmark datasets for 3D brain image registration and perform various experiments to evaluate our method. 
Experimental results show that our method clearly outperforms state-of-the-art methods on both benchmark datasets by a large margin.

\end{abstract}

\section{Introduction}

Image registration aims to transform different images into one system with the matched imaging contents, which has significant applications in brain image analysis, including brain atlas creation~\cite{chakravarty2006creation}, tumor growth monitoring~\cite{haskins2019deep} and multi-modality image fusion~\cite{du2016overview}. 
When we analyze a pair of brain images that were acquired from different sensors and viewpoints at different times, we need to transform one image (unaligned image) to another image (reference image) by establishing the anatomical correspondences~\cite{dalca2018unsupervised,fan2018adversarial,kuang2018faim}.
The correspondence between the unaligned image $x$ and the reference image $y$ is usually formulated by a transformation function $\phi_z$, which is parametrized by a latent variable $z$.

In order to calculate this latent variable, early works solved the optimization problems~\cite{avants2011reproducible,avants2008symmetric} in a high-dimensional deformation space, which is computationally expensive, thus limiting the practicability in clinical applications.
Recently, methods based on the deep convolutional neural networks (CNNs) learned the latent variable in an end-to-end manner, which largely reduces the computational time and shows results outperforming previous approaches.
For example, Sokooti~\emph{et al.}~\cite{sokooti2017nonrigid} developed a RegNet trained with the generated displacement vector fields to register CT images. 
Roh{\'e}~\emph{et al.}~\cite{rohe2017svf} learned to align the images by leveraging additional shape priors in a CNN.
However, these methods leverage the manually-labeled ground truth to train the deep networks in a supervised manner, where the labeled images are expensive and tedious to be obtained.
Hence, training on the limited labeled data degrades the performance of image registration. 
Very recently, researchers explored the unsupervised learning strategies to learn the transformation function between the unaligned image and moving image without ground truth labels.
Among them, Dalca~\emph{et al.}~\cite{dalca2018unsupervised} developed a probabilistic generative model for image registration by using CNN to learn the latent spatial transformation variable.
Krebs~\emph{et al.}~\cite{krebs2018unsupervised} applied conditional variable autoencoder to regularize low-dimensional probabilistic latent variables for image registration.
Kuang~\emph{et al.}~\cite{fan2018adversarial} employed different regularization choices in the deep network to predict the latent variable for a better registration result.
However, the existing deep-learning based methods take the unaligned and reference images as the input of a CNN and predict the latent variable directly, which \emph{ignores the transformation relationships between these two images in the feature levels}.
Thereby, the features learned at hidden layers of the CNN are not ``transparent'' to the latent variable, which reduces the discriminativeness of features for image registration.

In this paper, we present to \emph{introduce direct regularizations to the hidden layers} of two deep convolutional neural networks (CNNs): one CNN to extract features from the unaligned image while another CNN from the reference image. 
We provide the regularizations by \emph{adopting probabilistic models to capture the transformation relationship between each pair of hidden layers} in these two CNNs.
These probabilistic models can be seen as the additional constraints to regularize the intermediate feature maps during the learning process.
Furthermore, we embed the regularization terms into multiple layers of the CNNs and produce the feature-level latent variables in different layers.
Finally, we combine the predicted feature-level latent variables of all layers and predict the final latent variable for 3D brain image registration.
The whole network is trained end-to-end in an unsupervised manner. 
Experimental results on LPBA40 and MindBoggle101 dataset demonstrate that our method outperforms the current state-of-the-art methods by a large margin.
\section{Methodology}

\subsection{Method Overview}

\begin{figure}[t!]
    \centering
    \includegraphics[width=\textwidth]{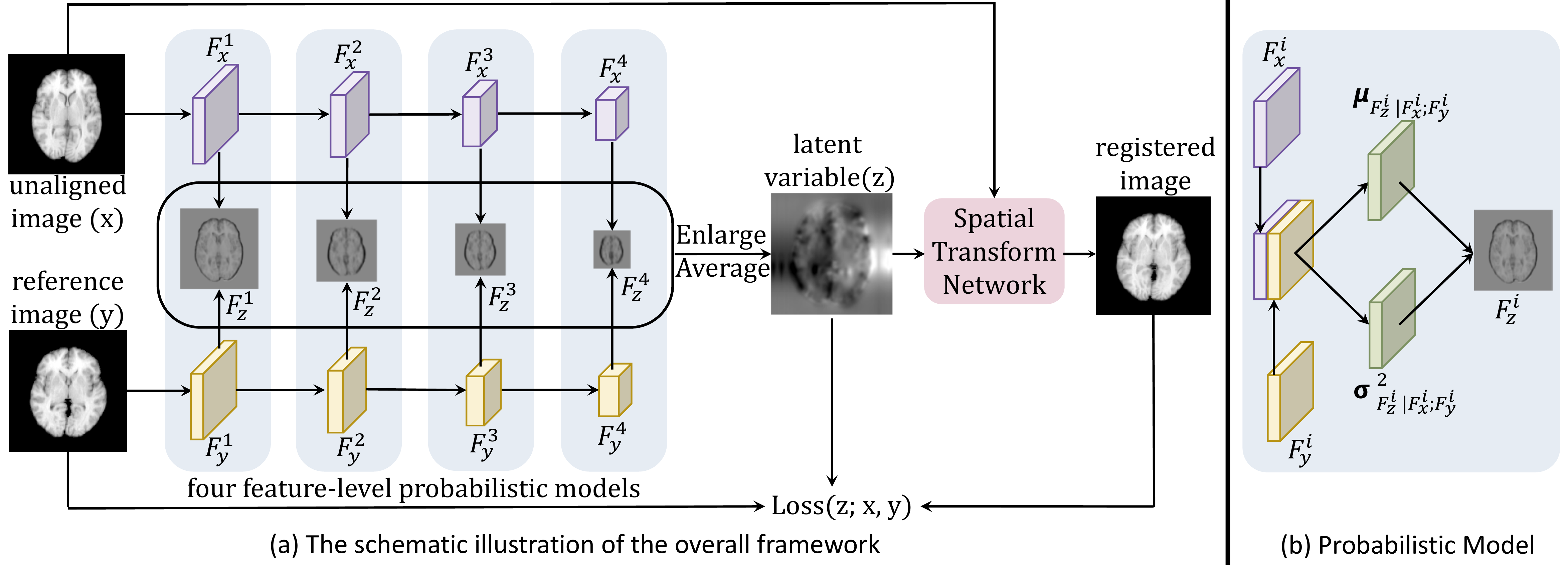}
    \vspace{-5.5mm}
    \caption{(a) The schematic illustration of the overall framework. (b) The feature-level probabilistic model used in each pair of feature maps.}
    \label{fig:fig_network}
    \vspace{-2.5mm}
\end{figure}

Figure~\ref{fig:fig_network} presents the workflow of the overall architecture of our proposed network for the 3D brain image registration.
To begin with, our method produces two sets of feature maps with different spatial resolutions by using two CNNs, which take the unaligned and reference 3D brain image (denoted as $x$ and $y$) as the inputs.
Then, we design a feature-level probabilistic inference model (see Section~\ref{probabilistic_model_model} and Figure~\ref{fig:fig_network}(b)) to estimate the feature-level latent variable, which represents the transformation relationship between the feature maps in the same layers of these two CNNs.
We carry this estimation from the top layer (with the highest spatial resolution) to the bottom layer (with the lowest spatial resolution) in the CNNs.
After that, we enlarge the estimated feature-level latent variables from all CNN layers to the same size, add them together to produce the final latent variable $z$.
%
Finally, we feed $x$ and $z$ into a spatial transform network (STN)~\cite{jaderberg2015spatial} to generate the aligned image.

\subsection{Feature-level Probabilistic Model} \label{probabilistic_model_model}

Given a pair of feature maps ($F_{x}^{i}$, $F_{y}^{i}$) from the $i$-th layer of the two CNNs, our probabilistic model aims to estimate the $i$-th latent varible $F_{z}^{i}$, which parametrizes a spatial transformation function (denoted as $\psi_{F_{z}^{i}}$) for mapping $F_{x}^{i}$ to $F_{y}^{i}$.
According to the probabilistic model, we estimate $F_{z}^{i}$ by maximizing the posterior registration probability $p(F_{z}^{i}|F_{x}^{i};F_{y}^{i})$ from the observed $F_{x}^{i}$ and $F_{y}^{i}$.
Similar to other works~\cite{dalca2018unsupervised,krebs2018unsupervised}, we adopt a variational approach to compute $p(F_{z}^{i}|F_{x}^{i};F_{y}^{i})$ by first introducing an approximate posterior probability $q_{\psi}(F_{z}^{i}|F_{x}^{i};F_{y}^{i})$ and then minimizing a KL divergence between $p(F_{z}^{i}|F_{x}^{i};F_{y}^{i})$ and $q_{\psi}(F_{z}^{i}|F_{x}^{i};F_{y}^{i})$ to make these two distributions as similar as possible.

The minimization of KL divergence between $p(F_{z}^{i}|F_{x}^{i};F_{y}^{i})$ and $q_{\psi}(F_{z}^{i}|F_{x}^{i};F_{y}^{i})$ is defined as:
\begin{equation} 
\begin{aligned} \label{Eq:KL-loss}
    &\min_{\psi}KL[q_{\psi}(F_{z}^{i}|F_{x}^{i};F_{y}^{i}) \ || \ p(F_{z}^{i}|F_{x}^{i};F_{y}^{i})] \\
  = &\min_{\psi}KL[q_{\psi}(F_{z}^{i}|F_{x}^{i};F_{y}^{i}) \ || \ p(F_{z}^{i})] - E_{q}log \ p(F_{y}^{i}|F_{z}^{i};F_{x}^{i}) \ , 
\end{aligned}
\end{equation}
where $q_{\psi}(F_{z}^{i}|F_{x}^{i};F_{y}^{i})$ comes from a multivariate normal distribution $\mathcal{N}$:
\begin{equation} \label{Eq:app-loss}
    q_{\psi}(F_{z}^{i}|F_{x}^{i};F_{y}^{i})=\mathcal{N}(z;\mu_{F_{z}^{i}|F_{x}^{i};F_{y}^{i}}, \sigma_{F_{z}^{i}|F_{x}^{i};F_{y}^{i}}^2) \ ,
\end{equation}
where $\mu_{F_{z}^{i}|F_{x}^{i};F_{y}^{i}}$ and $\sigma_{F_{z}^{i}|F_{x}^{i};F_{y}^{i}}$ are the mean and standard variance of the distribution, and they are directly learned through the convolutional layers (see Figure~\ref{fig:fig_network}(b)) by using the combined feature maps of $F_{x}^{i}$ and $F_{y}^{i}$. 
The $p(F_{z}^{i})$ and $p(F_{y}^{i}|F_{z}^{i};F_{x}^{i})$ follow the multivariate normal distribution, which are modeled as:
\begin{equation}
    p(F_{z}^{i}) = \mathcal{N}(F_{z}^{i};0,\sigma_{F_{z}^{i}}^2) \ , 
\end{equation}
\begin{equation}
    p(F_{y}^{i}|F_{z}^{i};F_{x}^{i}) = \mathcal{N}(F_{y}^{i};F_{x}^{i} \circ \phi_{F_{z}^{i}}, \sigma_{F^{i}}^2) \ , 
\end{equation}
where $\sigma_{F_{z}^{i}}$ is the variance (a diagonal matrix) of this distribution and $F_{x}^{i} \circ \phi_{F_{z}^{i}}$ is the noisy observed registered feature maps in which $\sigma_{F^{i}}^2$ is the variance of the noisy term; see~\cite{dalca2018unsupervised} for detail definition.

\subsection{Multilayer Fusion Network}

\begin{figure}[t!]
    \centering
    \includegraphics[width=\textwidth]{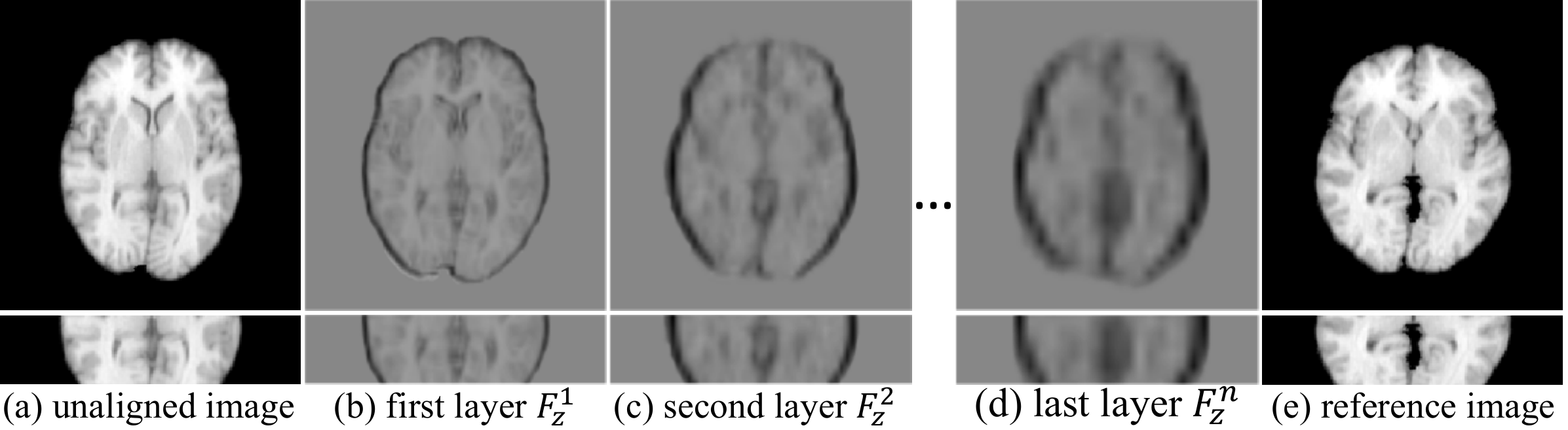}
    \vspace{-7.5mm}
    \caption{The visualization of (a) unaligned image; (b)-(d) the learned latent variables for different layers (from shallow layer to deep layer); (e) reference image.}
    \label{fig:fig_multilayer}
\end{figure}

As shown in Figure~\ref{fig:fig_multilayer}, the feature maps at shallow CNN layers have high resolutions but with fruitful detail information, while the feature
maps at deep layers have low resolutions but with high-level semantic information. 
The highly semantic features can help to register the global shape but neglecting many subtle details in the latent spatial variable; see Figure~\ref{fig:fig_multilayer}(d).
In contrast, as shown in Figure~\ref{fig:fig_multilayer}(b), the fruitful details in the low-level latent variable capture the local detail registration deformation but fail to generate global shape registration correspondence.
Motivated by this, we present to predict the final latent spatial transformation function between the two input images by leveraging features at multiple layers of the CNN to boost the registration performance.
To achieve this, we leverage the feature-level probabilistic model to estimate the feature-level latent variable $F_{z}^{i}$ by taking feature pairs from the shallowest layer to the deepest layer and fuse them together to compute the final latent variable $z$, which is then feed into the a spatial transform network (STN)~\cite{jaderberg2015spatial} for generating the output registered image of our network.

\subsection{Training and Testing Strategies}

\subsubsection{Loss function.}
Our network predicts the latent registration transformation variable at each CNN layer and the output registered image. The total loss (denoted as $\mathcal{D}_{total}$) for each pair of images is defined as  
\begin{equation} \label{Eq:total-loss}
    \mathcal{D}_{total}  = \mathcal{L}(z; x, y) + \sum_{i=1}^{n}  w_{i}\mathcal{L}(F_{z}^{i}; F_{x}^{i}, F_{y}^{i}) \ ,
\end{equation}
where $\mathcal{L}(z; x, y)$ denotes the KL divergence based loss of the prediction of the output registered image from the two input images (x and y); where $\mathcal{L}(F_{z}^{i}; F_{x}^{i}, F_{y}^{i})$ is the KL divergence based loss generated of the registration transformation variable ($F_{z}^{i}$) prediction by taking two feature maps ($F_{x}^{i}$ and ($F_{y}^{i}$)) at the $i$-th layer of the CNNs as the input.  $n$ is the number of CNN layer and $w_{i}$ is the loss weight of the $i$-th layer. We empirically set $n$ and $w_{i}$ as 4 and 1, respectively.
According to~\cite{dalca2018unsupervised}, we start from the KL divergence in Eq.~\ref{Eq:KL-loss} and define the KL divergence-based loss in Eq~\ref{Eq:total-loss} as:
\begin{equation} \label{Eq:real-KL-loss}
    \mathcal{L}(Z; X, Y) = \frac{1}{2 \sigma_{Z|X;Y}^2} || Y - X \circ \phi_{Z} ||^2 + \frac{1}{2}[tr(\sigma_{Z|X;Y}^2) + || \mu_{Z|X;Y} || - log \  det(\sigma_{Z|X;Y}^2)] \ ,
\end{equation}
where $Z \in [z, F_{z}^{i}]$, $X \in [x, F_{x}^{i}]$ and $Y \in [y, F_{y}^{i}]$.
The first term of Eq.~\ref{Eq:real-KL-loss} is a reconstruction loss for enforcing the registered image $X \circ \phi_{Z}$ to be similar to reference image $Y$.
The second term is a closed form of first term of Eq.~\ref{Eq:KL-loss}, and it encourages $q_{\psi}(Z|X;Y)$ and $p(Z)$ to be close.
$\mu_{Z|X;Y}$ and $\sigma_{Z|X;Y}$ are the mean and standard variance of the distribution $q_{\psi}(Z|X;Y)$, and they are directly learned through convolutional layers; see Eq.~\ref{Eq:app-loss}.

\vspace{-4mm}
\subsubsection{Training parameters.}

In our proposed model, we used the encoder architecture in~\cite{dalca2018unsupervised} as the backbone for both CNNs.
We adopted the initialization strategy of this work~\cite{he2015delving} to initialize the weights of all convolutional layers.
Moreover, we set the initial learning rate as $1e^{-4}$, periodically reduced it by multiplying with $0.1$, and stopped the learning process after $100$ epochs.
We employed the Adam optimizer~\cite{kingma2014adam} with the first momentum of 0.9, the second momentum of 0.999, and a weight decay of $0.0001$ to minimize the loss (see Eq.~\ref{Eq:total-loss}) of the whole network.
Our network was implemented using the Keras toolbox with a Tensorflow backend and we set the mini-batch size as one. 

\vspace{-4mm}
\subsubsection{Inference.}
Given an unaligned image and a reference image, our network first estimates a latent variable from these two images and produces the registered image of the unaligned image by feeding the unaligned image and latent variable into the spatial transformation network (STN).
Finally, we take the predicted registered image as the final output of our framework.
%
%
%
%

%
\section{Experiments}
%
%
%

\subsection{Benchmark Datasets and Evaluation Metric}

\subsubsection{LPBA40.}
The LONI Probabilistic Brain Atlas (LPBA40) dataset~\cite{shattuck2008construction} consists of 40 T1-weighted 3D brain MRI images from healthy subjects, and each volume has a brain mask and the corresponding segmentation mask (56 anatomical labels).
We used the first 30 volumes as the training data and the remaining 10 volumes as the testing data. 
Note that we didn't use any segmentation mask during the training process.

\vspace{-4mm}
\subsubsection{MindBoggle101.}

MindBoggle101 dataset~\cite{klein2012101} contains 101 skull-stripped T1-weighted 3D brain MRI images from healthy subjects, and only 62 MRI images have their segmentation masks.
For a fair comparison, we followed the recent work~\cite{kuang2018faim} and adopted 42 images for training and 20 images for testing.

%

\vspace{-4mm}
\subsubsection{Data preprocessing \& Evaluation metric.}
 
We conducted the preprocessing steps for each 3D brain image, where these steps include brain extraction, voxel spacing re-sampling (1mm), affine spatial normalization, ``Whitening'' operation, and intensity normalization; see~\cite{dalca2018unsupervised} for detail. To evaluate the registration performance, we register each unaligned image as well as its segmentation mask,
and measure the overlap between the registered segmentation mask and the segmentation mask of reference image using a widely-used Dice metric; see~\cite{dalca2018unsupervised,fan2018adversarial,kuang2018faim} for the details of the Dice definition.
In general, a larger $Dice$ indicates a better 3D brain registration result.

\subsection{Experimental Results}
\subsubsection{Quantitative comparison.}

\begin{table}[t]
\centering
\caption{Comparison with the state-of-the-arts using Dice on the LPBA40 dataset.}
\label{lpba40_results}
\resizebox{\textwidth}{!}{
    \begin{tabular}{l|p{1.5cm}<{\centering}|p{1.5cm}<{\centering}|p{1.5cm}<{\centering}|p{1.5cm}<{\centering}|p{1.5cm}<{\centering}|p{1.5cm}<{\centering}|p{1.5cm}<{\centering}} \hline \toprule[1pt]
                                            & Frontal               & Parietal              & Occipital             & Temporal              & Cingulate             & Putamen               & Hippo                 \\ \midrule[0.8pt]
    UtilzReg~\cite{vialard2012diffeomorphic}                                & 0.691                 & 0.617                 & 0.612                 & 0.665                 & 0.665                 & 0.710                 & 0.692                 \\ \hline
    VoxelMorph~\cite{dalca2018unsupervised}                              & 0.669                 & 0.610                 & 0.605                 & 0.652                 & 0.663                 & 0.700                 & 0.689                 \\ \hline
    FAIM~\cite{kuang2018faim}                                    & 0.676                 & 0.617                 & 0.608                 & 0.658                 & 0.675                 & 0.710                 & 0.696                 \\ \midrule[0.8pt]
    Baseline-1                              & 0.677                 & 0.637                 & 0.617                 & 0.635                 & 0.622                 & 0.549                 & 0.601                 \\ \hline
    Baseline-2                              & 0.702                 & 0.661                 & 0.648                 & 0.664                 & 0.665                 & 0.602                 & 0.682                 \\ \hline
    Our Method                              & \textbf{0.711}                 & \textbf{0.661}                 & \textbf{0.660}                 & \textbf{0.679}                 & \textbf{0.691}                 & \textbf{0.711}                 & \textbf{0.701}                 \\ \bottomrule[2pt]
    \end{tabular}
}
\end{table}

\begin{table}[t]
\centering
\caption{Comparison with the state-of-the-arts using Dice on the MindBoggle dataset.}
\label{MindBogglelpba40_results}
\resizebox{0.8\textwidth}{!}{
    \begin{tabular}{l|p{1.5cm}<{\centering}|p{1.5cm}<{\centering}|p{1.5cm}<{\centering}|p{1.5cm}<{\centering}|p{1.5cm}<{\centering}} \hline \toprule[1pt]
                                            & Frontal               & Parietal              & Occipital             & Temporal              & Cingulate          \\ \midrule[0.8pt]
    UtilzReg~\cite{vialard2012diffeomorphic}                                & 0.482                 & 0.456                 & 0.425                 & 0.385                 & 0.446              \\ \hline
    VoxelMorph~\cite{dalca2018unsupervised}                              & 0.534                 & 0.527                 & 0.510                 & 0.433                 & 0.483              \\ \hline
    FAIM~\cite{kuang2018faim}                                    & 0.572                 & 0.551                 & \textbf{0.537}        & 0.469                 & 0.508              \\ \midrule[0.8pt]
    Baseline-1                              & 0.502                 & 0.478                 & 0.448                 & 0.505                 & 0.536              \\ \hline
    Baseline-2                              & 0.560                 & 0.545                 & 0.433                 & 0.523                 & 0.546              \\ \hline
    Our Method                              & \textbf{0.579}        & \textbf{0.559}        & 0.430                 & \textbf{0.544}        & \textbf{0.546}                   \\ \bottomrule[2pt]
    \end{tabular}
}
\end{table}

We compared our method with three recent unsupervised brain image registration methods: UtilzReg~\cite{vialard2012diffeomorphic}, VoxelMorph~\cite{dalca2018unsupervised}, and FAIM~\cite{kuang2018faim}. Among them, UtilzReg adopted the hand-crafted features for image registration while the other two methods applied the CNN to predict the lantern variable for image registration.
For a fair comparison, we obtained their results either by directly taking
the results from their papers or by generating the results from the public codes provided by the authors using the recommended parameter setting.
Moreover, we followed~\cite{kuang2018faim} and reported the Dice value on seven large regions of the human brain on LPBA40 dataset, where these regions are obtained by grouping all the tissues according to the regions of interests; see~\cite{kuang2018faim} for the details.
For MindBoggle101 dataset, we followed~\cite{dalca2018unsupervised,kuang2018faim} and compared the results on five large cortical regions, which are grouped from $25$ cortical regions.

Table~\ref{lpba40_results} \&~\ref{MindBogglelpba40_results} summary the quantitative results in terms of Dice on unsupervised 3D brain image registration in the two benchmark datasets. 
Apparently, Our method outperforms all the others for almost all the cases on these two benchmark datasets. 
It demonstrates that by introducing direct regularizations to multiple hidden layers in the deep convolutional neural networks, we can obtain more discriminative features for latent variable prediction, thus producing more accurate registration results.

\vspace{-4mm}
\subsubsection{Visual comparison.}
\begin{figure}[t!]
    \centering
    \includegraphics[width=0.92\textwidth]{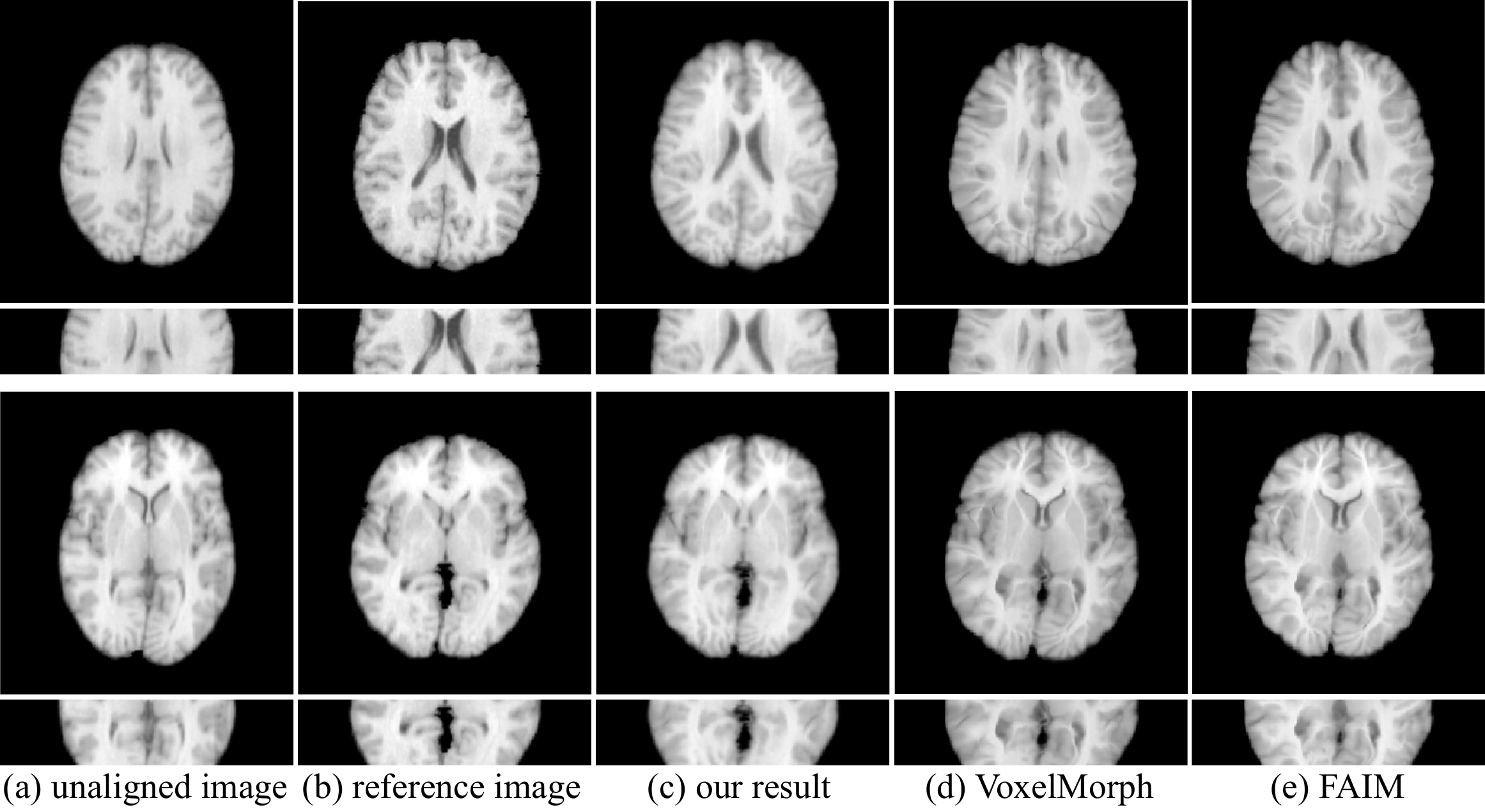}
    \vspace{-4.5mm}
    \caption{Visual comparison of the results produced by our method and other methods.}
    \label{fig:fig_multilayer}
    \vspace{-5mm}
\end{figure}

Figure~\ref{fig:fig_multilayer} presents the visual comparison results produced by different registration methods. 
From the results, we can see that other methods tend to fail to the match the shape of the reference image or lose the structure details while our method is able to produce the result that is more consistent with the reference image and better preserves the internal structures.  

\vspace{-4mm}
\subsubsection{Ablation study.}
We performed an ablation study to evaluate the major components in our network design.
Here, we considered two baselines. 
The first (denoted as ``Baseline-1'') was a framework constructed by replacing all the feature-level probabilistic models (Section~\ref{probabilistic_model_model}) with the concatenation of $F_{x}$ and $F_{y}$ between two networks shown in Figure~\ref{fig:fig_network};
the second (denoted as ``Baseline-2'') computed the latent variable only from the feature maps at the last CNN layers.
Table~\ref{lpba40_results} \&~\ref{MindBogglelpba40_results} reported the comparison results, showing that the designed probabilistic model can effectively capture
the transformation relationship between each pair of hidden layers in the two networks and adopting the probabilistic models to regularize multiple CNN layers leads to further improvement.

%
\section{Conclusion}
This paper presents a deep neural network for boosting the 3D brain image registration.
Our key idea is to develop feature-level probabilistic models to estimate the latent registration transformation variables from multiple layers of two convolutional neural networks (CNNs), which are constructed from two input images. 
Our network can provide the direct regularizations for hidden CNN layers and these direct regularizations introduce additional constraints for predicting the registration transformation variable, producing more discriminative features for image registration.
Experimental results on two benchmark datasets demonstrate that our network clearly outperforms state-of-the-art methods.
%

\subsubsection{Acknowledgment.} The work described in this paper was supported by a grant from the 
Research Grants Council of Hong Kong Special Administrative Region, China
(Project No. CUHK 14225616).

\vspace{-2.5mm}
{\small
    \bibliographystyle{bib/splncs03}
    \bibliography{bib/refs}
}

\end{document}